\documentclass[10pt,twocolumn,letterpaper]{article}

\usepackage{cvpr}
\usepackage{times}
\usepackage{epsfig}
\usepackage{graphicx}
\usepackage{amsmath}
\usepackage{amssymb}

\usepackage{multirow}

\newcommand*\samethanks[1][\value{footnote}]{\footnotemark[#1]}

\usepackage[pagebackref=true,breaklinks=true,letterpaper=true,colorlinks,bookmarks=false]{hyperref}
\cvprfinalcopy 

\def\cvprPaperID{0006} 

\ifcvprfinal\fi
\begin{document}

\title{Differential Morph Face Detection using Discriminative Wavelet Sub-bands}
\author{Baaria Chaudhary\thanks{Authors contributed equally.} , Poorya Aghdaie\samethanks[1] , Sobhan Soleymani, Jeremy Dawson, Nasser M. Nasrabadi \\
West Virginia University \\
{\tt\small \{bac0062,pa0002,ssoleyma\}@mix.wvu.edu, \{jeremy.dawson,nasser.nasrabadi\}@mail.wvu.edu }}

\pagenumbering{gobble}

\maketitle
\begin{abstract}
Face recognition systems are extremely vulnerable to morphing attacks, in which a morphed facial reference image can be successfully verified as two or more distinct identities. In this paper, we propose a morph attack detection algorithm that leverages an undecimated 2D Discrete Wavelet Transform (DWT) for identifying morphed face images. The core of our framework is that artifacts resulting from the morphing process that are not discernible in the image domain can be more easily identified in the spatial frequency domain. A discriminative wavelet sub-band can accentuate the disparity between a real and a morphed image. To this end, multi-level DWT is applied to all images, yielding 48 mid and high-frequency sub-bands each. The entropy distributions for each sub-band are calculated separately for both bona fide and morph images. For some of the sub-bands, there is a marked difference between the entropy of the sub-band in a bona fide image and the identical sub-band's entropy in a morphed image. Consequently, we employ Kullback-Liebler Divergence (KLD) to exploit these differences and isolate the sub-bands that are the most discriminative. We measure how discriminative a sub-band is by its KLD value and the  22 sub-bands with the highest KLD values are chosen for network training. Then, we train a deep Siamese neural network using these 22 selected sub-bands for differential morph attack detection. We examine the efficacy of discriminative wavelet sub-bands for morph attack detection and show that a deep neural network trained on these sub-bands can accurately identify morph imagery.
\end{abstract}

\begin{figure*}
\begin{center}
\includegraphics[scale=0.287]{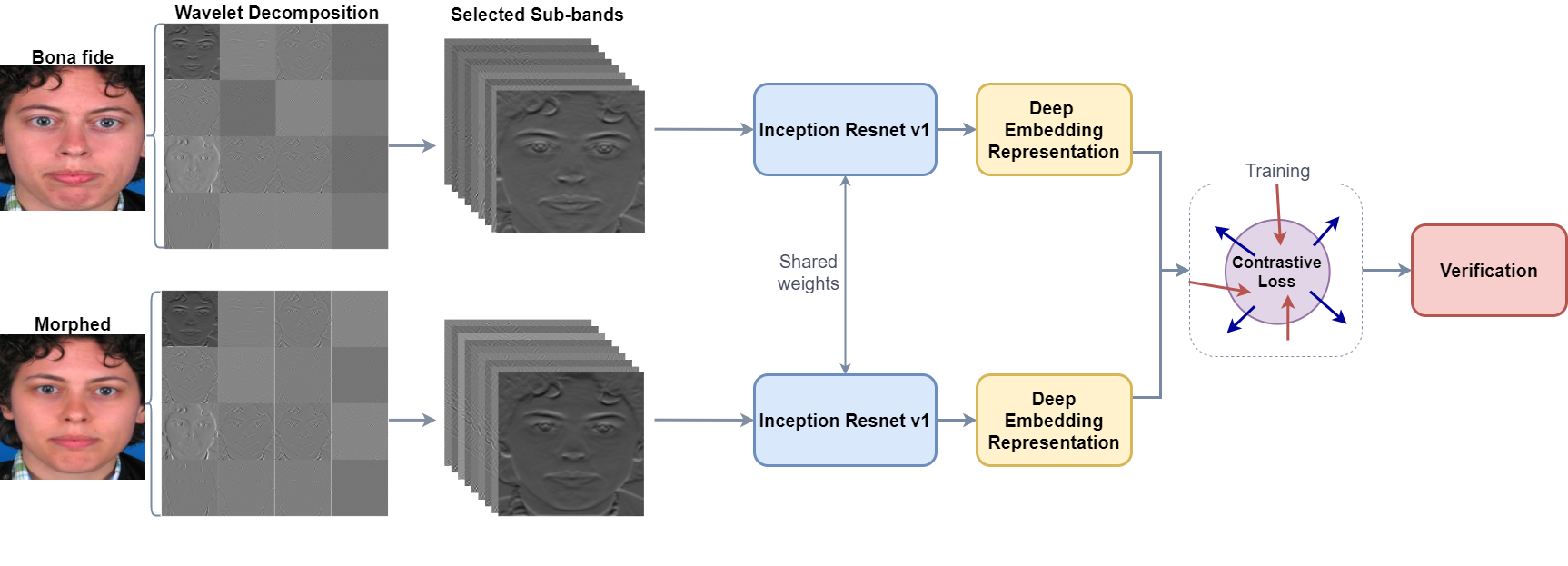}
\end{center}
   \caption{\textbf{Network architecture.} Image pairs are decomposed into wavelet sub-bands, only selecting the discriminative wavelet sub-bands for network training. The selected sub-bands are then sent to the Inception ResNET v1, where contrastive loss is applied during training. From the feature embedding representation, verification is performed by computing the $L_2$ distance between the feature vectors and a decision score is produced. }
\label{network_architecture}
\end{figure*}


\section{Introduction}

Face recognition systems are increasingly replacing human inspectors in border control and other security applications. Face capture is non-invasive, can be performed at a distance, and benefits from a relatively high social acceptance. Furthermore, face recognition systems also have a natural safeguard: if the algorithm triggers a false alarm, a human expert on site can easily perform the verification. For these reasons, the International Civil Aviation Organization (ICAO) has mandated the inclusion of a facial reference image in all passports worldwide \cite{icao2015machine}. Still, all the characteristics that make face recognition systems appealing also make them vulnerable. The mass adoption of automatic biometric systems for border control has exposed the inability of these systems to reliably detect a falsified image in a passport, particularly a morphed image, which has been identified as a serious threat. A morph attack is when a single morphed face image can be positively verified as two or more distinct identities \cite{ferrara2014magic}. This type of attack requires no complex forgery of passport technology, but rather a simple manipulation of the passport photo at time of application. Many morphing applications are easily and freely accessible and have no knowledge barrier \cite{mallick2016face}. It follows that a criminal attacker, who otherwise cannot travel freely, could obtain a legitimate travel document by morphing his face with that of an accomplice with similar features, resulting in existing face recognition systems verifying the morph image as either of the two individuals. 

Morphed images are not visually perceptible to the human eye, which makes them especially difficult to detect. As characteristics of both subjects are taken into account when morphing an image, face recognition systems are easily deceived and, as such, the false acceptance rate is very high. In addition, these systems are designed to tolerate a large intra-class variance to account for the significant changes in facial appearance that occur in the 5 to 10-year life cycle of a passport. Many commercial off-the-shelf (COTS) systems have repeatedly failed to detect morphed images \cite{robertson2017fraudulent}. Similarly, studies also show human recognizers are unable to correctly differentiate between a morphed image and an authentic one \cite{robertson2017fraudulent} \cite{raghavendra2017face} \cite{bourlai2016face}. Even after instruction on how to detect a morphed image, human recognizers still perform worse than face recognition algorithms \cite{makrushinvisapp19} \cite{kramer2019face}.

In this paper, we propose a differential morph attack detection algorithm using an undecimated 2D Discrete Wavelet Transform (DWT). By decomposing an image to wavelet sub-bands, we can identify the morph artifacts that are hidden in the image domain more clearly in the spatial frequency domain. Our analysis of wavelet sub-bands demonstrates that specific high-frequency components are more discriminative for morph attack detection. To this end, multi-level DWT is applied to the images, yielding 48 mid- and high-frequency sub-bands each and discarding the low-frequency bands altogether. To isolate the most informative sub-bands, we employ Kullback-Liebler Divergence (KLD) \cite{kullback1951information} on the sub-band entropy distributions. The higher the KLD value, the more discriminative the sub-band is for morph detection. We then use the selected informative sub-bands to train a deep Siamese network for the differential morph attack detection scenario. The Siamese network takes bona fide and morph pairs as input and yields a confidence score on the likelihood of the pairs being from the same person as shown in Figure \ref{network_architecture}. Siamese networks are ideal for this scheme as they are primarily employed in tasks that require finding similarities between two inputs. We examine the usefulness of wavelet sub-bands for differential morph attack detection and show a deep neural network trained on these sub-bands can accurately identify morph imagery. The experiments are conducted on three different morph image datasets: VISAPP \cite{visapp17}, MorGAN \cite{damer2018morgan} \cite{debiasi2019detection}, and LMA \cite{damer2018morgan}. Additionally, cross-dataset performance is evaluated on AMSL \cite{neubert2018extended}.

The paper is organized as follows: we briefly summarize the related works in Section \ref{relatedwork}, explain the methodology in the Section \ref{method}, and discuss our experiments and subsequent results in Section \ref{experiments}. Finally, conclusions are presented in Section \ref{conclusion}.







\section{Related Work}\label{relatedwork}

The vulnerability of face recognition systems to morph attacks was first introduced by \cite{ferrara2014magic}. Many morph detection algorithms have been proposed since then for both single (no reference) and differential morph attack detection scenarios. Single (no reference) morph attack detection algorithms rely only on the potential morphed image to make their classification. Conversely, differential morph attack detection algorithms compare the potential morphed image with an additional trusted image, typically a live capture at border security. As such, differential morph attack detection algorithms have more information at their disposal to make their classification and, therefore, generally perform better than single morph detection algorithms \cite{scherhag2018towards}. 

With respect to single morph attack detection, many classical hand-crafted feature extraction techniques have been explored. The most well-performing of these general image descriptors is Binarized Statistical Image Features (BSIF) \cite{kannala2012bsif}, in which extracted BSIF features were classified using a Support Vector Machine (SVM) \cite{raghavendra2016detecting}. However, deep learning methods consistently perform better than these general feature extraction techniques \cite{seibold2017detection} \cite{wandzik2017cnns} \cite{raghavendra2017transferable}. In \cite{raghavendra2017transferable}, complementary features from pre-trained VGG-19 and AlexNet models are concatenated and then used to train a Probabilistic Collaborative Representation-based Classifier (ProCRC).  The authors of \cite{scherhag2018fusion} employ a multi-algorithm fusion approach by extracting feature vectors using BSIF, LBP \cite{liao2007learning}, SIFT \cite{lowe2004sift}, SURF \cite{bay2006surf} and HOG with additional deep feature embeddings from OpenFace DNN \cite{amos2016openface} used as the last feature vector. These feature vectors are then used to train separate SVMs, applying score-level fusion at the end to obtain the final decision score. Photo Response Non-Uniformity (PRNU) spectral analysis has also been proposed to analyze the alterations caused by morphing features \cite{scherhag2019PRNU}. In \cite{neubert2019face}, the authors design a face morphing detector by combining spatial and frequency feature descriptors from an image. Fuzzy LBP in color channels of HSV and YCbCR color spaces are investigated in \cite{ramachandra2019towards}. Additionally, studying the residual noise computed on color channels using deep CNN-based denoising has also been presented for reliable face morphing detection \cite{venkatesh2019morphed} \cite{venkatesh2020detecting}. This paper aggregates several denoised instances of an image in the wavelet domain. 

There are a few papers that also address differential morph attack detection. Face demorphing has provided some encouraging results \cite{ferrara2017face} \cite{ferrara2018face}. In simplistic terms, the demorphing algorithm subtracts the potential morph image from the trusted image.  The difference image is then compared to the trusted image and a low similarity score signifies a morphed image. Face demorphing has also been approached using a GAN framework to restore an accomplice's image \cite{peng2018fdgan}.  Classical feature extraction methods have also been applied to the differential scenario by taking the difference of the feature vectors of the images being compared. This difference vector along with the original feature vector for the potential morph is then used to train a difference SVM and a feature SVM, respectively. This method is explored in \cite{scherhag2018towards}, using LBP, BSIF, SIFT, SURF, and HOG descriptors. Scherhag et al. \cite{scherhag2020deep} uses deep face representations from feature embeddings extracted from ArcFace \cite{deng2019arcface} to detect a morph attack. The authors also emphasized the need for high variance and trained their network on a morph database constructed using multiple different morph generation techniques. Disentanglement of appearance and landmarks is another method recently proposed for differential morph detection \cite{soleymani2020differential}. The use of Siamese networks for differential morph image detection has also been explored \cite{Soleymani_2021_WACV}, but only in the image domain.

Because of the lack of large, publicly available morph database, many morph detection algorithms train on small morph datasets, usually created in house. However, researchers can submit their morph detection algorithms to the NIST FRVT morph detection test \cite{ngan2020face} for objective evaluation. Most of the algorithms submitted to NIST exhibit less than ideal performance on almost all tested morph datasets, which vary in quality and method. The deep learning method in \cite{scherhag2020deep} outperforms the other models in the NIST test, most likely due to the training protocol employing cross-database training.
\begin{figure}[t]
\begin{center}
\includegraphics[scale=0.2715]{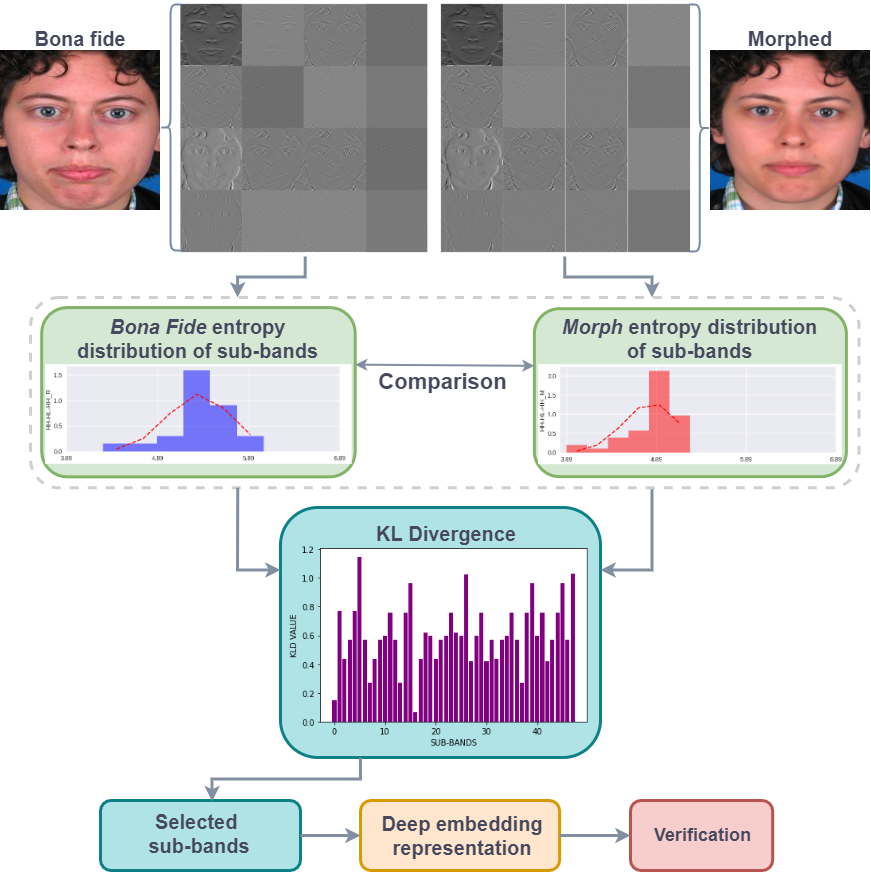}
\rule{0.1\linewidth}{0pt}
\end{center}
  \caption{\textbf{Discriminative wavelet sub-band selection algorithm.} Bona fide (left) and morphed (right) images are decomposed into 48 wavelet sub-bands each. The entropy distributions and corresponding KL divergence values are found. For a given sub-band, the dissimilarity between the bona fide and morph entropy distribution represents how informative the sub-band is for morph detection. KL divergence is applied to isolate the discriminative sub-bands. A Siamese network is trained with the selected informative sub-bands.}
  \label{wavelet_band_selection}
\end{figure}

\section{Method}\label{method}
\begin{figure}[t]
\begin{center}
\includegraphics[scale=0.164]{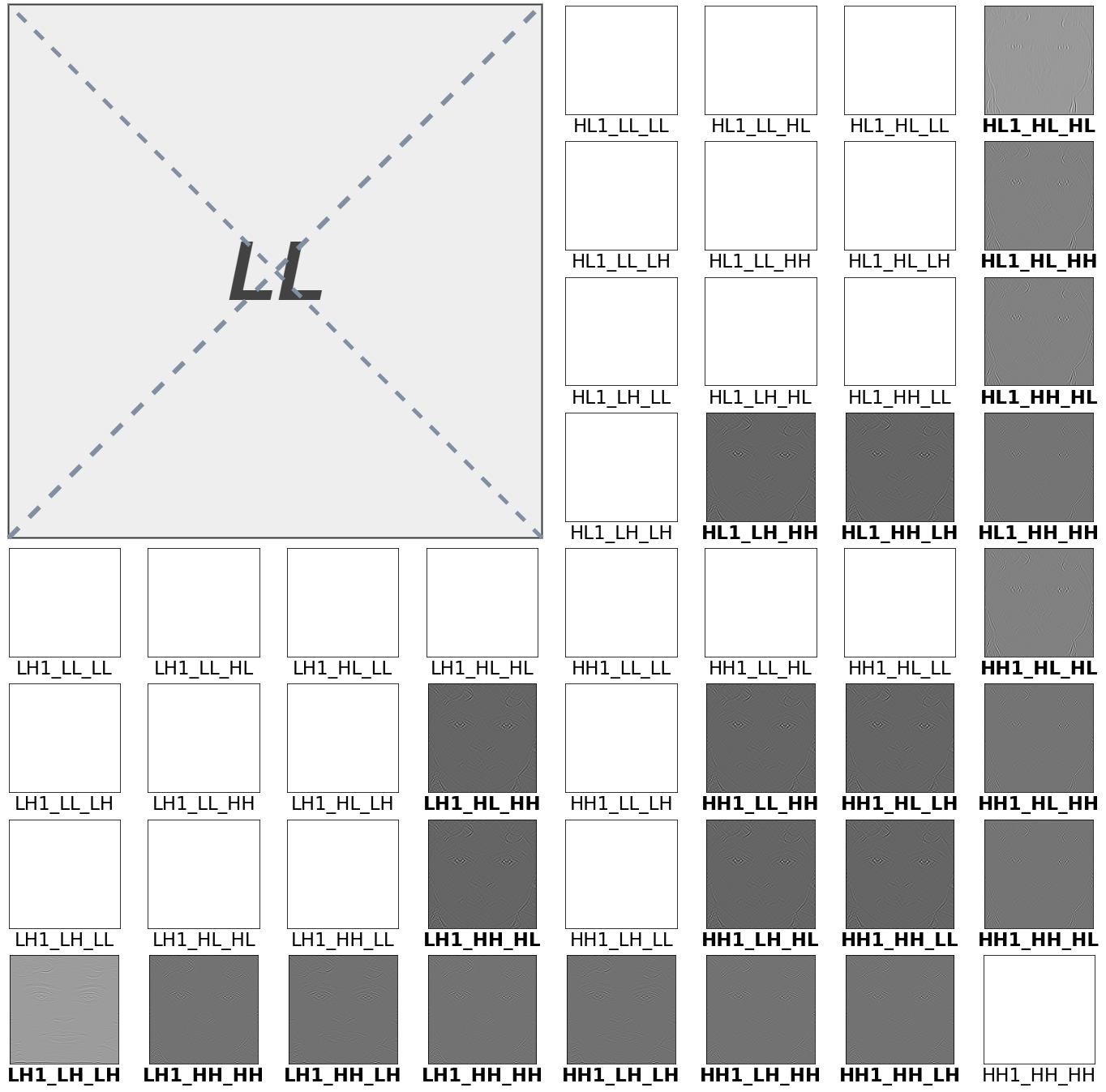}
\rule{0.1\linewidth}{0pt}
\end{center}
  \caption{\textbf{Selected sub-bands.} The selected sub-bands are shown with regards to their location in wavelet decomposition. Most of the informative sub-bands chosen by KL divergence are those that have been filtered with the HH filter.}
  \label{subbands_grid}
\end{figure}

Our morph attack detection framework centers around applying undecimated 2D wavelet decomposition and training a Siamese deep neural network to classify morphs based on the most discriminative wavelet sub-bands. Because the mrophing process can involuntarily introduce artifacts in the final morph image, the proposed method aims to isolate these artifacts in the wavelet domain and effectively utilize them for morph detection. A close study of the wavelet sub-bands shows that most morphing artifacts reside in the high frequency spectrum. As such, we do not consider the Low Low (LL) sub-band for decomposition and drop the LL sub-band completely after the first level of wavelet decomposition. Instead, we decompose only the Low High (LH), High Low (HL) and High High (HH) sub-bands down to the third level. After three levels of uniform decomposition, 48 sub-bands are obtained per image. We determine the most optimal sub-bands for network training using Shannon entropy \cite{shannon1948mathematical} and KL divergence \cite{kullback1951information}. After instituting a threshold over the KLD values of the 48 sub-bands, we obtain 22 sub-bands with the highest KLD values. Figure \ref{subbands_grid} displays the sub-bands that are selected for network training and their location in the wavelet decomposition. A final set of 22 informative sub-bands is then used to train a Siamese deep neural network, consisting of the Inception ResNET v1 architecture as the base network. 

\subsection{2D Discrete Wavelet Transform}
A 2D Wavelet transform decomposes an image in the frequency domain, essentially capturing different frequencies at different resolutions. This means that wavelet transform allows us to separately examine the approximation and detail data in an image. Particularly for morph detection, we can pinpoint the sub-bands where the morph artifacts appear and discard the sub-bands that are not informative for our problem. 

Wavelet decomposition occurs by applying the low-pass and high-pass filters both vertically and horizontally simultaneously on a given image. After one level of decomposition, the LL, LH, HL, and HH sub-bands are obtained. We can continue decomposing the image further by filtering each sub-band separately. In our framework, we adopt undecimated wavelet decomposition, which maintains the resolution of the image with each decomposition, and decompose the LH, HL, and HH sub-bands specifically down to the third level. Our chosen naming convention for the sub-bands is such that each sub-band is labeled LH\_HL\_HH, where LH is the sub-band after the first decomposition, HL is the sub-band after the second decomposition, and HH is the sub-band after the third level of decomposition. As morphed images are, in essence, approximations of the original, our research indicates that the LL sub-band is unhelpful for morph detection.


 \begin{figure*}[ht]
\begin{center}
\includegraphics[scale=0.335]{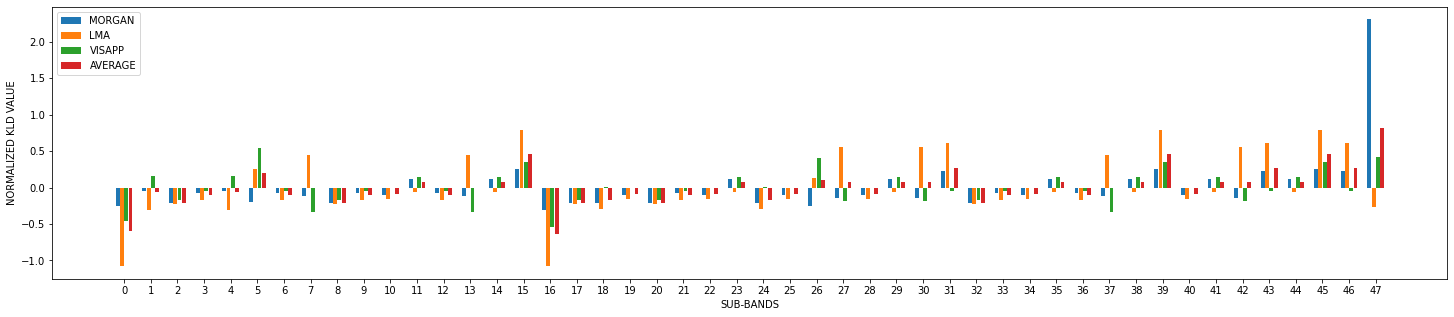}
\rule{0.1\linewidth}{0pt}
\end{center}
   \caption{Normalized KL divergence values in the sub-bands 0 to 48 for all three morphing techniques: LMA (orange), MorGAN (blue), and VISAPP (green). The averaged KL divergence value is represented in red.}
   \label{KLD}
\end{figure*}

\subsection{Sub-band Selection using KL Divergence} \label{bandselection}

Training on all 48 sub-bands does not isolate which sub-bands truly contribute to the classification result. Consequently, we employ Shannon entropy and Kullback-Liebler divergence to identify the optimal sub-bands for morph attack detection. Shannon entropy, in particular, is used to measure the embedded information in each sub-band. Figure \ref{wavelet_band_selection} illustrates the wavelet sub-band selection algorithm.  For each of the three datasets, Shannon entropy and the entropy distributions are computed for all 48 sub-bands. Since we are interested in the comparison of bona fide and morphed images, we calculate the entropy distributions for all bona fide and morphed images in a dataset separately. Then, KL divergence (relative entropy) is calculated between the bona fide entropy distribution and the morph entropy distribution for each sub-band. 

The method for finding the KLD values is as follows:  after the entropy distributions of each sub-band are found, we find the histograms of entropy of all 48 sub-bands for both morphed and bona fide images. Accordingly, 96 normal distributions (48 bona fide and 48 morphed) are estimated using these histograms. ${\hat{f}}_{b_i}$ represents the estimated bona fide normal distribution for the $i^{th}$ sub-band, and similarly, ${\hat{f}}_{m_i}$ represents the estimated morph normal distribution for the $i^{th}$ sub-band. Dissimilarity of the two probability distribution functions, namely $({\hat{f}}_{b_i},{\hat{f}}_{m_i})$ is calculated for all 48 sub-bands and the KL divergence is computed for each relative entropy distribution.  

The KLD values vary by dataset as each dataset is created using a different morphing technique. Therefore, we focus on selecting the sub-bands that are discriminative across different morphing techniques. As such, the KL divergence values of each dataset are normalized by removing the mean. Then, the normalized values are averaged over the KLD values of each sub-band for each of the three datasets. Figure \ref{KLD} presents the distribution of the normalized KL divergence values for the three morphed datasets and their average values. The higher the normalized KLD value for a single sub-band, the more informative the sub-band is for morph classification. After sorting the normalized KLD values from highest to lowest, we institute a threshold for selecting the sub-bands for training.  According to the method described in \cite{pooryaIWBF}, the optimal number of sub-bands is found to be 22. Thus, our final network is trained with 22 input channels, consisting of the top-22 most discriminative sub-bands for morph detection. 


    


    

\subsection{Siamese Network}
A Siamese neural network \cite{bromley1994signature} is the architecture used to train with the wavelet sub-bands. A Siamese network consists of two identical sub-networks which share weights. Siamese networks are ideal for morph attack detection as they are primarily designed to find similarities between two inputs. Contrastive loss \cite{hadsell2006dimensionality} is the loss function utilized for training the Siamese network. Contrastive loss is a distance-based loss function which attempts to bring similar images closer together in a common latent space. At the same time, the loss function distances the dissimilar ones even more. Essentially, contrastive loss seeks to emphasize the similarity between samples of the same class and exaggerate the differences between images of different classes. The distance is found from the feature embeddings of the input pair produced by the Siamese network. The margin is the distance threshold that regulates the extent to which pairs are separated. The equation for calculating contrastive loss is as follows: 

\begin{equation}
L_c=(1-y_g)D(I_1,I_2)^2+y_g\max(0,m-D(I_1,I_2))^2,
\label{eq:1}
\end{equation}
where $I_1$ and $I_2$ are the input face images, $m$ is the margin or distance threshold to control the separation and $y_g$ is the ground truth label for a given pair of training images and $D(I_1,I_2)$ is the $L_2$ distance between the feature vectors:

\begin{equation}
D(I_1,I_2)=||\phi(I_1)-\phi(I_2)||_2.
\label{eq:2}
\end{equation}

Here, $\phi(.)$ represents a non-linear deep network mapping image into a vector representation in the embedding space. According to the loss function defined above, $y_g$ is $0$ for genuine image pairs and $y_g$ is $1$ for imposter (morph) pairs. 


\begin{table*}[t]
\footnotesize
    \centering
    \caption{Performance of the proposed framework and baselines. With the exception of RGB-66 testing on MorGAN, BW-22 exhibits superior performance.}
    \begin{tabular}{|l|c|c|c|c|c|c|}
    \hline
    \multirow{2}{*}{Testing}&\multirow{2}{*}{Method}&\multicolumn{2}{c}{{APCER@BPCER}}&\multicolumn{2}{c}{{BPCER@APCER}}&D-EER\\& &5\%&	10\%	&5\%&	10\%& \%\\ 
    \hline
    \multirow{6}{*}{MorGAN} &BW images &7.88&6.17&13.1&3.1&5.57\\
                        &RGB images&4.5&3.3&3.22&1.74&4.17\\
                        &LL-removed BW images&5.5&3.14&4.5&3.28&5.53\\
                        &LL-removed RGB images&3.66&2.98&1.58&0.79&3.55\\
                        &BW-22 wavelets&3.71&1.85&3.06&0.26&3.89\\
                        & \textbf{RGB-66 wavelets}&\textbf{0.86}&\textbf{0.0}&\textbf{0.37}&\textbf{0.37}&\textbf{1.62}\\
    \hline
    \multirow{6}{*}{LMA} &BW images &22.7&14.3&36.5&15.1&11.6\\
                        &RGB images&11.1&6.68&12.2&5.62&8.8\\
                        &LL-removed BW images&25.9&14.4&19.0&11.5&11.5\\
                        &LL-removed RGB images&15.75&7.4&12&6.48&8.06\\
                        &\textbf{BW-22 wavelets}&\textbf{4.95}&\textbf{2.67}&\textbf{4.38}&\textbf{1.46}&\textbf{4.52}\\
                        &RGB-66 wavelets&10.53&5.39&9.44&4.72&7.36\\
    \hline
    \multirow{6}{*}{VISAPP} &BW images &5.97&0.0&0.0&0.0&3.17\\
                        &RGB images&1.32&0.08&0.0&0.0&0.0\\
                        &LL-removed BW images&1.57&0.08&5.63&4.22&0.0\\
                        &LL-removed RGB images&2.98&0.8&0.0&0.0&3.25\\
                        &\textbf{BW-22 wavelets}&\textbf{0.0}&\textbf{0.0}&\textbf{0.0}&\textbf{0.0}&\textbf{0.0}\\
                        &RGB-66 wavelets&0.0&0.0&0.0&0.0&0.0\\
    \hline
    \multirow{6}{*}{UNIVERSAL} &BW images &15.0&8.95&14.4&7.5&8.53\\
                        &RGB images&6.65&4.01&5.22&2.5&5.63\\
                        &LL-removed BW images&19.1&6.74&10.872&7.78&8.45\\
                        &LL-removed RGB images&10.9&3.53&5.52&4.56&5.52\\
                        &\textbf{BW-22 wavelets}&\textbf{3.25}&\textbf{1.69}&\textbf{3.01}&\textbf{0.65}&\textbf{3.93}\\
                        &RGB-66 wavelets&6.4&2.67&5.15&2.57&5.15\\
    \hline
    \end{tabular}
    
    \label{tab:table}
\end{table*}

To streamline training, a Siamese Inception ResNET v1 architecture \cite{szegedy2017inception} is adopted, using weights pre-trained on the VGGFace2 dataset \cite{cao2018vggface2}. The network is then re-trained with the morphing datasets for the differential Siamese implementation. The model is optimized by enforcing contrastive loss on the embedding space representations of the genuine and imposter morph samples. The pre-trained Siamese network is then additionally fine-tuned using the training portion of each morph database. The feature embeddings are taken from the last fully connected layer and the $L_2$ distance between the two embeddings is calculated for verification.

\section{Experiments}\label{experiments}

\subsection{Datasets}
We train our network on three different morph datasets that apply three different morphing techniques: splicing, GAN generation, and landmark manipulation to investigate how our model generalizes. The two morph image databases used in this experiment are VISAPP \cite{visapp17} and MorGAN \cite{damer2018morgan} \cite{debiasi2019detection}.  VISAPP is a collection of complete and splicing morphs generated using the Utrecht FCVP database \cite{Utrecht}. The images are $900\times 1200$ in size. This dataset is generated by warping and alpha-blending two face images together \cite{wolberg1998image} and then splicing the resulting face into one of the faces of the original contributing images. This preserves the background and hairline of one of the contributing faces, which helps avoid blurry artifacts and ghosting that typically occurs in these regions and makes morphs easier to spot \cite{visapp17}. For our network, we only use a subset of 183 high quality splicing morphs that is constructed by selecting the morph images that have no recognizable artifacts (VISAPP-Splicing-Selected dataset) along with 131 genuine neutral and smiling images for a total of 314 images.

The MorGAN database is generated from a selection of full frontal face images manually chosen from the CelebA dataset \cite{liu2018large}. It consists of a custom morph image generation pipeline (MorGAN) \cite{damer2018morgan}, created by the authors that uses a GAN, inspired by the learned inference model \cite{dumoulin2016adversarially}, to generate morphs. The encoder in the GAN transforms the images into a latent space and when two latent spaces related to two different subjects are combined, a morphed image is synthesized. The database consists of 1,500 bona fide reference images, 1,500 bona fide probe images and 1,000 MorGAN morphs of size $64\times64$ pixels. To compare their GAN morphs, the authors also generate 1,000 LMA (landmark manipulation) morphs \cite{mallick2016face}. The VISAPP and MorGAN images differ significantly in terms of quality and resolution. However, varying quality and resolution during training can result in a network that is more robust and performs better on different morphing techniques, particularly when the bona fide images are of equal resolution to the morphed images.  An additional publicly available dataset, the Advanced Multimedia Security Lab (AMSL) Face Morph Image dataset \cite{neubert2018extended}, is used as an ``unseen''' dataset to measure cross database performance.  This dataset consists of approximately 102 bona fide images and 2,175 morph images, created using the Combined Morph tool, as described in \cite{neubert2018extended}.  This dataset is used for testing purposes only.

All images are preprocessed according to the FaceNet architecture \cite{schroff2015facenet}: face detection and alignment is performed via MTCNN \cite{zhang2016joint}. All images are resized to $160\times160$ pixels before uniform wavelet decomposition is applied. According to the KLD sub-band selection algorithm, the top 22 wavelet sub-bands are selected for each image to prepare for network training. As the Siamese network expects pairs of input, the morph wavelet bands are paired off into genuine face pairs and imposter face pairs, where a genuine pair consists of two trusted 22 selected wavelet sub-bands and an imposter pair consists of a trusted image's wavelet sub-bands and a corresponding morph image's wavelet sub-bands. 50\% of the subjects are considered for training while the other 50\% are used to evaluate the performance of the network. In addition, 15\% of the test set is selected during model optimization as the validation set. The training data is further augmented with horizontal flips to increase the training set and improve generalization. By design, the train-test split is disjoint, with no overlapping morphs or contributing bona fides to morphs. This enables us to attain an accurate representation of performance. Batch size of 64 pairs of 22 selected wavelet sub-bands of size $22\times160\times160$ is used for training the model. The batch generator also compensates for class imbalance, ensuring that the network sees an equal number of morph pairs and genuine pairs every iteration.

\subsection{Network Setup and Metrics}
We fine-tune an Inception ResNET v1, already pretrained on VGGFace2, on the 22 selected sub-bands. We train the network using the training portions of all three datasets, calling it the ``universal'' dataset. The margin \textit{m} of contrastive loss is set to 1. Adam is the chosen optimizer and the initial learning rate is 0.0001. The performance is monitored by the validation loss and whenever the validation loss achieves a new low, the best weights are saved. Every time the validation loss plateaus, the best weights are re-loaded, the learning rate is divided by 10, and training continues from there down to 1e-07. After that, early stopping is implemented if the loss still does not improve after 35 epochs. The network is implemented using PyTorch and training is accelerated with the use of three 12 GB Titan X (Pascal) GPUs. We train the network on the universal dataset for 150 epochs. All experimental results in the paper are reported for the final iteration.

The standard quantitative measures for morph attacks are used to measure performance: APCER and BPCER. Attack Presentation Classification Error Rate (APCER) is the percentage of morphed samples incorrectly classified as bona fide. Conversely, the Bona-fide Presentation Classification Error Rate (BPCER) is the percentage of bona fide images classified as morphs. D-EER stands for Detection Equal Error Rate at which APCER equals BPCER. The APCER5 is the APCER rate when BPCER = 5\% and similarly APCER10 is the APCER rate when BPCER is 10\%. In real world applications, the BPCER rate is the measure by which individuals are inconvenienced with a false alarm. We plot these rates in a Detection Error Tradeoff (DET) graph.
\begin{figure}[t]
\begin{center}
\includegraphics[scale=0.518]{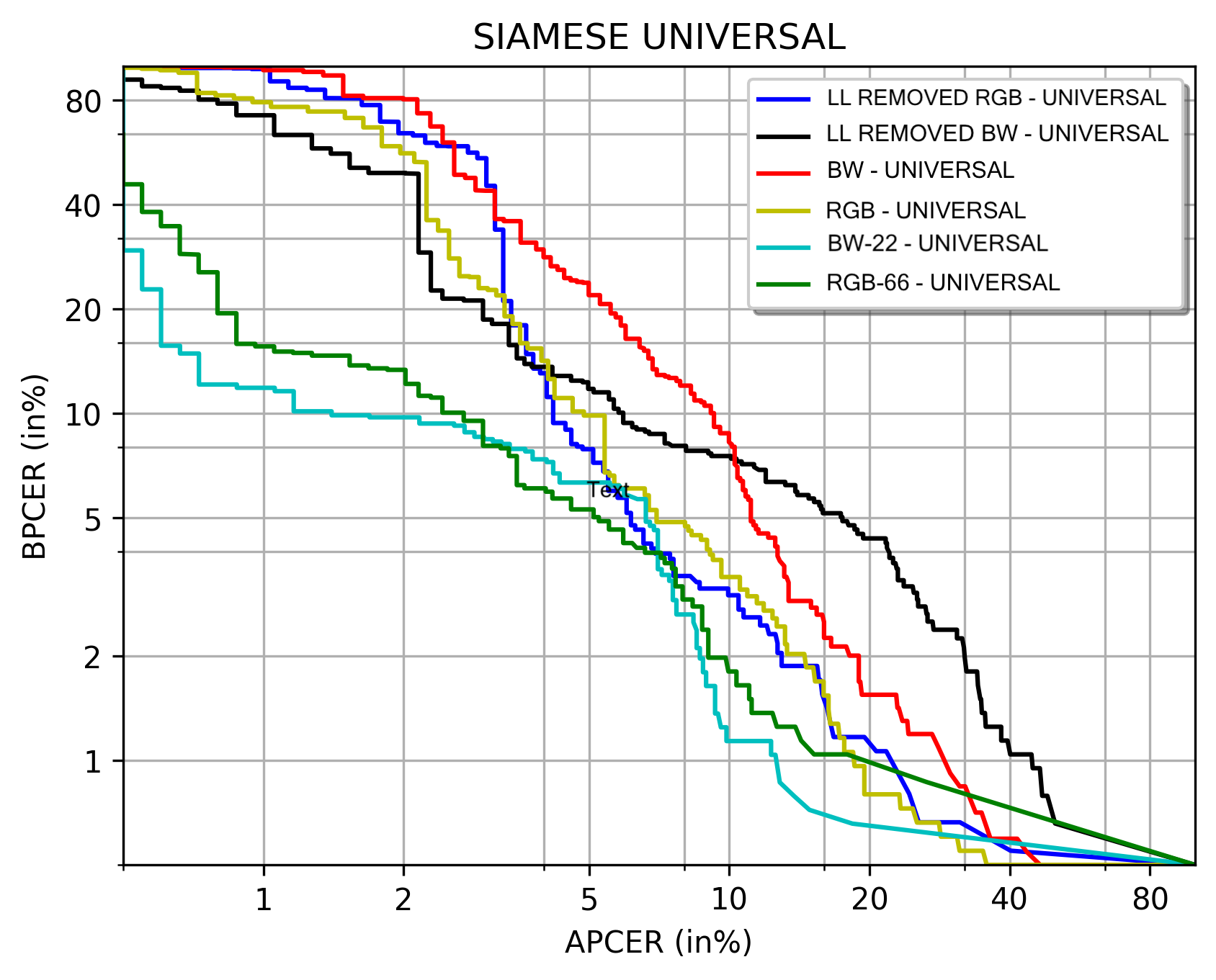} 

\end{center}
    \caption{DET curves for all protocols, tested on the universal test set. The universal test set consists of the testing portions of all three datasets and serves as an indicator for overall network performance.}
   \label{DET}
\end{figure}
\subsection{Results}
We assess the performance of our morph detector using the test data of each individual dataset as well as the universal test set (which consists of all three individual test sets). The universal test set performs essentially as an ``average''  of how the universally trained network performs on each of the individual databases. We  feel it is important to capture how the network learns each morphing technique separately and generally.  As is standard for wavelet transform, the images are converted to grayscale before wavelet decomposition is applied. The 22 most discriminative sub-bands are selected for training according to the procedure described in Section \ref{bandselection}. These grayscale 22 selected sub-bands, called BW-22, are then used to train a Siamese network for morph detection. 

To determine how wavelets perform for RGB images, we also apply wavelet decomposition separately to each channel of an RGB image following the methodology described in Section \ref{method}, yielding 144 wavelet sub-bands (48 sub-bands per channel). 22 sub-bands are chosen for each channel, totaling 66 sub-bands each (RGB-66). The 22 sub-bands selected from each channel are the same as BW-22 to facilitate comparison. The training protocol is identical for both BW-22 and RGB-66, except for the batch size. Because RGB-66 consists of 66 channels of $160\times160$, the batch size is halved to 32 in our configuration to conserve memory.

Accordingly, we compare the performance of our wavelet Siamese networks, BW-22 and RGB-66, with other frameworks to validate the efficacy of our method. We apply each baseline to both color and grayscale images to gauge how color information plays a role in morph detection. The first baseline is RGB images (referred to as RGB) to compare how the original images' performance varies from that in the wavelet domain. We also train a Siamese network for the grayscale (referred to as BW) images to act as a baseline for BW-22 before wavelet decomposition is applied. Additionally, as the LL sub-band has shown to be unhelpful for morph detection, we have only decomposed the mid- and high-frequency information for our wavelet Siamese network. To mirror the removal of the approximation data in the image domain, we decompose the images into wavelet sub-bands. Then, we remove the LL-band in the wavelet domain, and reconstruct the image using Inverse Wavelet Transform. We do this for both RGB images and grayscale images in our dataset, designating these baselines as LL-removed RGB and LL-removed BW respectively. Theoretically, the LL-removed RGB and LL-removed BW baselines should be approximately equivalent to the original 144 sub-bands of RGB-66 and the 48 sub-bands of BW-22 before sub-band selection.  We train a separate Siamese network using the same training protocol for each of the above scenarios: RGB, BW, LL-removed RGB, LL-removed BW, RGB-66, and BW-22. 

\begin{figure}[t]
\begin{center}
\includegraphics[scale=0.518]{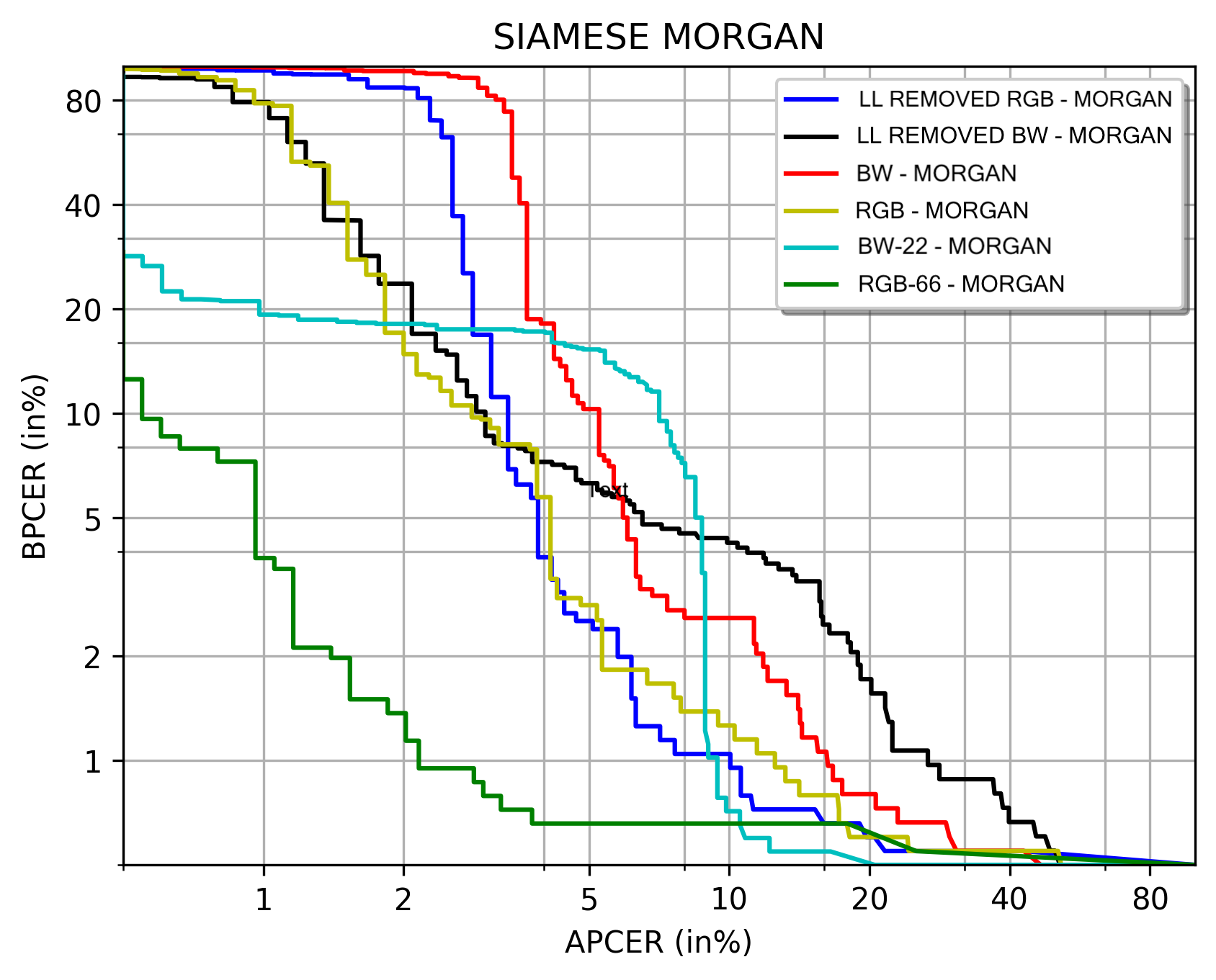} 

\end{center}
    \caption{DET curves for all protocols, tested on the MorGAN test set. RGB-66 performs the best for this morphing technique.}
   \label{DET_MORGAN}
\end{figure}

\begin{figure}[t]
\begin{center}
\includegraphics[scale=0.518]{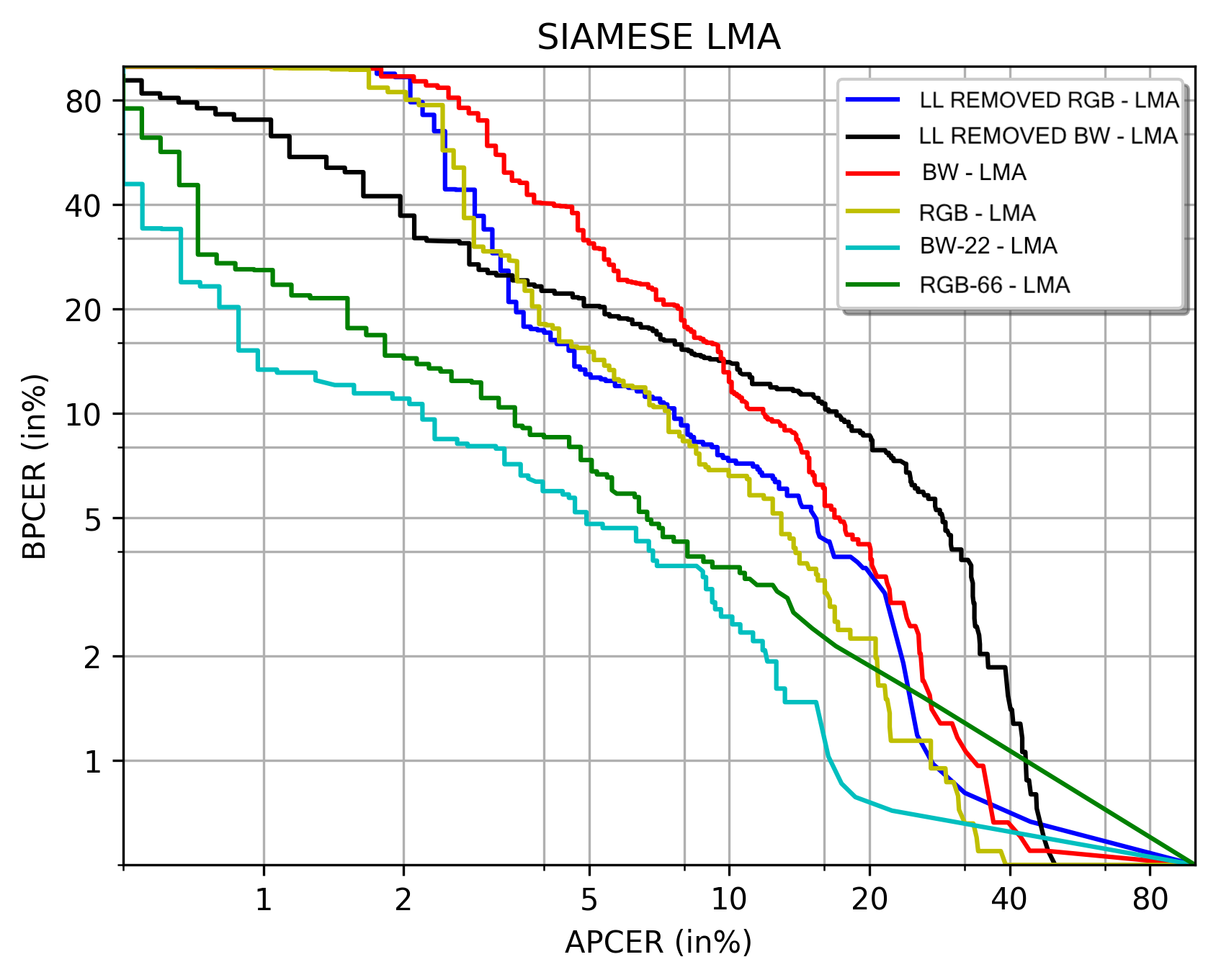} 

\end{center}
    \caption{DET curves for all protocols, tested on the LMA test set.}
   \label{DET_LMA}
\end{figure}

All networks are trained using the combined portions of the three datasets: VISAPP, LMA, and MorGAN to obtain a robust network that generalizes to many different morphing techniques. The performance of each network can be observed in Table \ref{tab:table}. From Table \ref{tab:table}, it is clear VISAPP performs consistently, regardless of framework. This is likely due to the small size of the VISAPP dataset that the models easily learn VISAPP's morphing technique. Generally speaking, the networks trained on RGB information perform better. This implies that there is in fact information in the color channels that is useful for morph detection. Notably, grayscale images exhibit poor performance all around. Table \ref{tab:table} also shows how difficult LMA morphs are to detect in comparison to MorGAN and VISAPP. This is in line with research that GAN-generated morphs are easier to detect than landmark manipulation morphs \cite{venkatesh2020can}. BW-22 performs significantly better on LMA morphs than the other models. From Figure \ref{DET_MORGAN}, it should be noted RGB-66 performs extremely well on the MorGAN dataset. This suggests that the MorGAN morphing technique contains more color information than LMA and VISAPP, which were creating using landmark-based techniques. MorGAN, on the other hand, was created using a GAN architecture, meaning it is essentially a synthesized image created using the two contributing images. Still, while RGB-66 performs unusually well for the MorGAN test set, BW-22 performs better on the LMA dataset (see Figure \ref{DET_LMA}) as well as overall performance as can be seen in Figure \ref{DET}, which shows the performance of the networks on the 'universal' dataset. From the results, we can derive that color information plays a smaller role in the classification of morphs in the wavelet domain. Figure \ref{DET} shows the DET curves for all tested networks. It is clear from Figure \ref{DET} that the wavelet Siamese networks, particularly BW-22, exhibit superior performance.

Additionally, in Table \ref{tab:table2}, we compare the performance of our wavelet-based morph detector with other classical feature extraction techniques that have been used for morph detection, namely BSIF \cite{kannala2012bsif}, LBP \cite{liao2007learning}, SIFT \cite{lowe2004sift}, and SURF \cite{bay2006surf}, each combined with an SVM. Each method is trained using the universal dataset and evaluated on the test sets of the individual datasets. We also measure the baseline performance of FaceNet \cite{schroff2015facenet} on all the morph test sets. To evaluate cross-database performance, AMSL dataset is used for testing only.

\begin{table}[t]
\footnotesize
    \centering
     \caption{Performance Comparison of Proposed Framework. All algorithms trained with the Universal dataset.}
    \begin{tabular}{|l|c|c|c|c|}
    \hline
    
    \multirow{2}{*}{Testing}&\multirow{2}{*}{Method}&\multicolumn{2}{c}{{APCER@BPCER}}&D-EER\\& &5\%&	10\%	& \%\\ 
    \hline
    \multirow{6}{*}{MorGAN} &SURF&86.8&70.11&46.1\\
                        &SIFT&57.6&47.7&27.3\\
                        &LBP&90.13&82.2&41.6\\
                        &BSIF&86.8&71.6&31.7\\
                        &FaceNet&36.80&31.15&22.25\\
                        & \textbf{BW-22 wavelets}&\textbf{0.86}&\textbf{0.0}&\textbf{1.62}\\
    \hline
    \multirow{6}{*}{LMA} &SURF &81.1&63.69&51.1\\
                        &SIFT&63.2&55.8&36.7\\
                        &LBP&91.1&83.4&40.5\\
                        &BSIF&86.5&75.0&36.4\\
                        &FaceNet&43.70&40.90&30.35\\
                        &\textbf{BW-22 wavelets}&\textbf{4.95}&\textbf{2.67}&\textbf{4.52}\\
    \hline
    \multirow{6}{*}{VISAPP} &SURF &94.1&90.3&47.8\\
                        &SIFT&91.1&84.7&52.2\\
                        &LBP&31.1&19.5&16.0\\
                        &BSIF&30.6&22.73&16.4\\
                        &FaceNet&25.0&15.8&15.5\\
                        &\textbf{BW-22 wavelets}&\textbf{0.0}&\textbf{0.0}&\textbf{0.0}\\
    \hline
    \multirow{6}{*}{AMSL} &SURF &96.7&91.3&53.0\\
                        &SIFT&94.65&84.9&38.0\\
                        &LBP&91.0&72.9&43.0\\
                        &BSIF&91.0&82.0&41.3\\
                        &FaceNet&38.6&31.35&19.86\\
                        &\textbf{BW-22 wavelets}&\textbf{33.78}&\textbf{23.61}&\textbf{16.4}\\
    \hline
    \end{tabular}
   
    \label{tab:table2}
\end{table}

\section{Conclusion} \label{conclusion}

In this paper, we introduced a framework to detect morphed face images using undecimated DWT. The core of our method was the ability to identify morph artifacts in the wavelet domain and to leverage the most informative sub-bands for differential morph detection. To select the optimal sub-bands, a data driven approach based on KL divergence is employed. The 22 selected sub-bands were then used to train a deep Siamese network successfully. Our framework achieves an EER of 3.93\% on the universal test set, significantly better than the other baselines. Furthermore, the framework performs well on an unseen morph dataset, AMSL, that uses a different morphing technique than our training set, achieving an EER of 16.4\%. This shows how wavelet decomposition with selective sub-bands is useful in the morph problem domain.

{\small
\bibliographystyle{ieee_fullname}
\bibliography{egpaper_for_review}
}

\end{document}